%% file: main.tex
\documentclass[runningheads]{llncs}
\usepackage{graphicx}
\usepackage{algorithm}
\usepackage{amsmath}
\usepackage{multirow}
\usepackage{amsfonts}

\begin{document}
\title{Exploiting the Matching Information in the Support Set for Few Shot Event Classification}
%
\titlerunning{Matching Information in the Support Set for Few Shot Event Classification}


\author{Viet Dac Lai\inst{1} \and
Franck Dernoncourt\inst{2} \and
Thien Huu Nguyen\inst{1}}


\authorrunning{V. D. Lai et al.}


%

\institute{Department of Computer and Information Science, University of Oregon, USA \and
Adobe Research, USA\\
\email{vietl@cs.uoregon.edu}, \email{dernonco@adobe.com}, \email{thien@cs.uoregon.edu}}

\maketitle       
\begin{abstract}
The existing event classification (EC) work primarily focuses on the traditional supervised learning setting in which models are unable to extract event mentions of new/unseen event types. Few-shot learning has not been investigated in this area although it enables EC models to extend their operation to unobserved event types. To fill in this gap, in this work, we investigate event classification under the few-shot learning setting. We propose a novel training method for this problem that extensively exploit the support set during the training process of a few-shot learning model. In particular, in addition to matching the query example with those in the support set for training, we seek to further match the examples within the support set themselves. This method provides more training signals for the models and can be applied to every metric-learning-based few-shot learning methods. Our extensive experiments on two benchmark EC datasets show that the proposed method can improve the best reported few-shot learning models by up to 10\% on accuracy for event classification.

\keywords{Event classification \and Auxiliary Loss \and Few-shot learning.}
\end{abstract}

\section{Introduction}

Event Classification (EC) is an important task of Information Extraction (IE) in Natural Language Processing (NLP). The target of EC is to classify the event mentions for some set of event types (i.e., classes). Event mentions are often associated with some words/phrases that are responsible to trigger the corresponding events in the sentences. For example, consider the following two sentences:

(1) \textit{The companies \textbf{fire} the employee who wrote anti-diversity memo}.

(2) \textit{The troops were ordered to cease \textbf{fire}}

In these examples, an EC system should be able to classify the word ``{\it fire}'' in the two above sentences as an {\it Employment-Termination} event and an {\it Attack} event, respectively. As demonstrated by the examples, a notable challenge in EC is that the similar surface forms of the words might convey different events depending on the context. Two main methods have been employed for EC. The first approach explores linguistic features (e.g., syntactic and semantic properties) to train statistical models \cite{Ji:08}. The second approach, on the other hand, focuses on developing deep neural network models (e.g., convolutional neural network (CNN) and recurrent neural network (RNN)) to automatically learn effective features from large scale datasets \cite{Chen:15,Liu:17}. Due to the development of the deep learning models, the performance for EC has been improved significantly \cite{Nguyen:15:event,Nguyen:16a:joint,Nguyen:16e,Lu:18,Nguyen:19}.

The current EC models mainly employ the traditional supervised learning setting \cite{Nguyen:15:event,Nguyen:16a:joint} where the set of event types for classification has been pre-determined. However, once a model is trained on the datasets with the given set of event types, it is unable to detect event mentions of unseen event types. To extend EC to new event types, a common solution is to annotate additional training data for such new event types and re-train the models, which is extremely expensive. It is thus desirable to formalize EC in the few-shot learning setting where the systems need to learn to recognize event mentions for new event types from a handful of examples. This is, in fact, closer to how humans learn to do tasks and make the EC models more applicable in practice. However, to our knowledge, there has been no prior work on few-shot learning for EC. 
 
In few-shot learning, we are given a support set and a query instance. The support set contains examples from a set of classes (e.g. events in EC). A learning model needs to predict the class, to which the query instance belongs, among the classes presented in the support set. This is done based on the matching information between the query example and those in the support set. To apply this setting to extract the examples of some new type, we need to collect just a few examples of the new type and add them to the support set to form a new class. Afterward, whenever we need to predict whether a new example has the new type or not, we can set it as the query example and perform the models in this setting. 

In practice, we often have some existing datasets (denoted by $D$) with examples for some pre-defined types. The previous work on few-shot learning has thus exploited such datasets to simulate the aforementioned few-shot learning setting to train the models \cite{snell2017prototypical}. Basically, in each episode of the training process, a subset of the types in $D$ is sampled for which a few examples are selected for each type to serve as the support set. Some other examples are also chosen from the remaining examples of each sampled type to establish the query points. The models would then be trained to correctly map the query examples to their corresponding types in the support set based on the context matching of the examples \cite{gao2019hybrid}.  

One potential issue with this training procedure is that the training signals for the models only come from the matching information between the query examples and the examples in the support set. The available matching information between the examples in the support set themselves is not yet explored in the existing few-shot learning work \cite{vinyals2016matching,snell2017prototypical}, especially for the NLP tasks \cite{gao2019hybrid}. While this approach can be acceptable for the tasks in computer vision, it might not be desirable for NLP applications, especially for EC. Overall, datasets in NLP are much smaller than those in computer vision, thus limiting the variety of the context for training purposes. The ignorance of the matching information for the examples in the support set might cause inefficiency in using the training data for EC where the models cannot fully exploit the available information and fail to achieve good performance. Consequently, in this work, we propose to simultaneously exploit the matching information between the examples in the support set and between the query examples with the examples in the support set to train the few-shot learning models for EC. This is done by adding additional terms in the loss function (i.e., the auxiliary losses) to capture the matching knowledge between the examples in the support set. We expect that this new training technique can better utilize the training data and improve the performance of few-shot learning in EC.

We extensively apply the proposed training method on different metric learning models for few-shot learning on two benchmark EC datasets. The experiments show that the new training technique can significantly improve all the considered few-shot learning methods over the two datasets with a large performance gap. In summary, the contribution of this work includes: (i) for the first time in the literature, we study the few-shot learning problem for event Classification, (ii) we propose a novel training technique for the few-shot learning models based on metric learning. The proposed training method exploits the matching information between the examples in the support set as additional training signals, and (iii) we achieve the state-of-the-art performance for EC on the few-shot learning setting, functioning as the baselines for the future research in this area.

\section{Related Work}

Early studies in event classification mainly focus on designing linguistic features \cite{Ahn:06,Ji:08,Li:14} for statistical models. Due to the development of deep learning, many advanced network architectures have been investigated to advance the event classification accuracy \cite{Chen:15,Nguyen:15:event,Nguyen:16a:joint,Nguyen:16b,Nguyen:16d,Liu:17,Nguyen:18a}. However, none of them investigates the few-shot learning problem for EC as we do in this work. Although some recent studies have considered a related setting where event types are augmented with some keywords \cite{bronstein:15:seed,Peng:16:minimal,Lai:19}, these works do not explicitly examine the few-shot learning setting as we do in this work. Some other efforts on zero-shot learning for event classification \cite{huang:18:zeroshot} are also related to our work in this paper.

Few-shot learning facilitates the models to learn effective latent features without large scale data. The early studies apply transfer learning to fine-tune the pre-trained models, exploiting the latent information from the common classes with adequate instances \cite{caruana1995learning,bengio2012deep}. Metric learning, on the other hand, learns to model the distance distribution among the observed classes \cite{koch2015siamese,vinyals2016matching,snell2017prototypical}. Recently, the idea of a fast learner that can generalize to a new concept quickly is introduced in meta-learning \cite{santoro2016meta,finn2017model}. Among these methods, metric-learning is more explainable and easier to train and implement compared to transfer learning and meta-learning. Notably, the prototypical networks in metric learning achieve state-of-the-art performance on several FSL benchmarks and show its robustness against noisy data \cite{snell2017prototypical,gao2019hybrid}. Although many FSL methods are proposed for image recognition \cite{koch2015siamese,vinyals2016matching,snell2017prototypical,finn2017model,santoro2016meta}, there have been few studies investigating this setting for NLP problems \cite{gao2019hybrid,yu2018diverse}.

\section{Methodology}
\subsection{Notation}

The task of few-shot event classification is to predict the event type of a query example $x$ given a support set $S$ and a set of event type $T = \{t_1, t_2, \ldots,t_N\}$ ($N$ is the number of event types). In few-shot learning, $S$ contains a few examples for each event type in $T$. For convenience, we denote the support set as: 
\begin{equation}
    \begin{aligned}
    S = & \{ (s^1_1, a^1_1, t_1),\ldots, (s^{K_1}_1, a^{K_1}_1, t_1)  \\ 
        &\ldots \\
        & ( s^1_N, a^1_N, t_N),\ldots, ( s^{K_N}_N, a^{K_N}_N, t_N)\},\\
    \end{aligned}
\end{equation}
where $(s^j_i, a^j_i, t_i)$ indicates that the $a^j_i$-th word in the sentence $s^j_i$ is the trigger word of an event mention with the event type $t_i$, and $K_1, K_2, \ldots, K_N$ are the numbers of examples in the support set for each type $t_1, t_2, \ldots, t_N$ respectively. For simplicity, we use $w_1, w_2, \ldots, w_l$ to represent the word sequence for some sentence with length $l$ in this work.

Similarly, the query example $x$ can also be represented by $x = (q,p,t)$ where $q$, $p$ and $t$ represent the query sentence, the position of the trigger word in the sentence, and the true event type for this event mention respectively. Note that $t \in T$ is only provided in the training time and the models need to predict this event type in the test time.

In practice, the numbers of support examples in $S$ (i.e., $K_1, \ldots, K_N$) may vary. However, to ease the processing and speed up the training process with GPU, similar to recent studies in FSL \cite{gao2019hybrid}, we employ the N-way K-shot FSL setting. In this setting, the numbers of instances per class in the support set are equal ($K_1 = \ldots= K_N=K > 1$) and small ($K\in\{5,10\}$).


Note that to evaluate the few-shot learning models for EC, we would need the training data $D_{train}$ and the test data $D_{test}$. For few-shot learning, it is crucial that the sets of event types in $D_{train}$ and $D_{test}$ are disjoint. The event type set $T$ in each episode would then be a sample of the sets of event types in $D_{train}$ or $D_{test}$, depending on the training and evaluation time respectively. Also, as mentioned in the introduction, in one episode of the training process, a set of query examples (i.e., the query set) would be sampled so it involves the similar event types $T$ as the support set, and the examples for each type in the query set would be different from those in the support set. At the test time, the classification accuracy of the models over all the examples in the test set would be evaluated.

\subsection{Few-shot Learning for Event Classification}

The few-shot learning framework for EC in this work follows the typical metric learning structures in the prototypical networks \cite{snell2017prototypical,gao2019hybrid}, involving three major components: instance encoder, prototypical module, classifier module.

\subsubsection{Instance encoder} 

Given a sentence $s=\{w_1, w_2, \ldots, w_l\}$ and the position of the trigger word $a$ (i.e., $w_a$ is the trigger word of the event mention in $s$ and $(s,a)$ can belong to an example in $S$ or the query example), following the common practice in EC \cite{Nguyen:15:event,Chen:15}, we first convert each word $w_i \in s$ into a real-valued vector to facilitate the neural computation in the following steps. In particular, in this work, we represent each word $w_i$ using the concatenation of the following two vectors:

\begin{itemize}
    \item The pre-trained word embedding of $w_i$: this vector is expected to capture the hidden syntactic and semantic information for $w_i$ \cite{Mikolov:13}.
    \item The position embedding of $w_i$: this vector is obtained by mapping its relative distance to the trigger word $w_a$ (i.e., $i-a$) to an embedding vector in the position embedding table. The position embedding table is initialized randomly and updated during the training process of the models. The purpose of the position embedding vectors is to explicitly inform the models of the position of the trigger word in the sentence \cite{Chen:15}.
\end{itemize}

After converting $w_i$ into a representation vector $e_i$, the input sentence $s$ becomes a sequence of representation vectors $E = e_1, e_2, \ldots, e_l$. Based on this sequence of vectors, a neural network architecture $f$ would be used to transform $E$ into an overall representation vector $v$ to encode the input example $(s,m)$ (i.e., $v = f(s,m)$). In this work, we investigate two network architectures for the encoding function $f$, i.e., one early architecture for EC based on CNN and one recent popular architecture for NLP based on Transformers:

 \textbf{CNN encoder}: This model applies the temporal convolution operation with some window size $k$ and multiple filters over the input vector sequence $E$, producing a hidden vector for each position in the input sentence. Such hidden vectors are then aggregated via the max-pooling operation to obtain the overall representation vector $v$ for $(s,m)$ \cite{Chen:15,gao2019hybrid}.

\textbf{Transformer encoder}: This is an advanced model to encode sequences of vectors based on attention mechanism without recurrent neural network \cite{vaswani2017attention}. The transformer encoder involves multiple layers; each of them consumes the sequence of hidden vectors from the previous layer to generate the sequence of hidden vectors for the current layer. The first layer would take $E$ as the input while the hidden vector sequence returned by the last layer (i.e., the vector at the position $a$ of the trigger word) would be used to constitute the overall representation vector $v$ in this case. Each layer in the transformer encoder is composed of two sublayers (i.e., a multi-head self-attention layer and a feed-forward layer) augmented with a residual connection around them \cite{vaswani2017attention}.

\subsubsection{Prototypical module}

The prototypical module aims to compute a single prototype vector to represent each class in $T$ of the support set. In this work, we consider two versions of this prototypical module in the literature. The first version is from the original prototypical networks \cite{snell2017prototypical}. It simply obtains the prototype vector $c_i$ for a class $t_i$ using the average of the representation vectors of the examples with the event type $t_i$ in the support set $S$:
\begin{equation}
\textbf{c}_i = \frac{1}{K}\sum_{(s^j_i, a^j_i, t_i) \in S} f(s^j_i, a^j_i)
\label{eq:c1}
\end{equation}

The second version, on the other hand, comes from the hybrid attention-based prototypical networks \cite{gao2019hybrid}. The prototype vector is a weighted sum of the representation vectors of the examples in the support set. The example weights (i.e., the attention weights) are determined by the similarity of the examples in the support set with respect to the query example $x = (q,p,t)$:
\begin{equation}
    \begin{aligned}
    \textbf{c}_i &= \sum_{(s^j_i, a^j_i, t_i)\in S} \alpha_{ij} f(s^j_i, a^j_i) \\
    \text{ where }  \alpha_{ij}&=\frac{\text{exp}(b_{ij})}{\sum_{(s^k_i, a^k_i, t_i)\in S} \text{exp}(b_{ik})}\\
    b_{ij} &= \sigma(f(s^j_i, a^j_i) \odot f(q,p))
    \label{eq:c2}
    \end{aligned}
\end{equation}

In this formula, $\odot$ is the element-wise multiplication and {\it sum} is the summation operation done over all the dimensions of the input vector.

\subsubsection{Classifier module}

In this module, we compute the probability distribution over the possible types for $x$ in $T$ using the distances from the query example $x=(q,p,t)$ to the prototypes of the classes/event types $T$ in the support set:
\begin{equation}
P(y=t_i|(q,p),S) = \frac{\text{exp}(-d(f(q,p), \textbf{c}_i))}{\sum^N_{j=1}\text{exp}(-d(f(q,p), \textbf{c}_j))}
\end{equation}
where $d$ is a distance function, and $\textbf{c}^i$ and $\textbf{c}^j$ are the prototype vectors obtained in either Equation (\ref{eq:c1}) or Equation (\ref{eq:c2}).

In this paper, we consider three popular distance functions in different few-shot learning models using metric learning:
\begin{itemize}
    \item Cosine similarity in matching networks (called \textbf{Matching}) \cite{vinyals2016matching}
    \item Euclidean distance in the prototypical networks. Depending on whether the prototype vectors are computed with Equation \ref{eq:c1} or \ref{eq:c2}, we have two variations of this distance function, called as \textbf{Proto} \cite{snell2017prototypical}, and \textbf{Proto+Att} (i.e., in hybrid attention-based prototypical networks \cite{gao2019hybrid}) respectively.
    \item Learnable distance function using convolutional neural networks in relation networks (called \textbf{Relation}) \cite{sung2018relation}
\end{itemize}

Given the probability distribution $P(y|x,S)$, the typical way to train the few shot learning framework is to optimize the negative log-likelihood function for $x$ (with $t$ as the ground-truth event type for $x$) \cite{snell2017prototypical,gao2019hybrid}:
\begin{equation}
    L_{query}(x, S) = - \log P(y = t|x,S)
    \label{eq:l1}
\end{equation}

\subsubsection{Matching the examples in the support set}

The typical loss function for few-shot learning in Equation \ref{eq:l1} aims to learn by matching the query example $x$ with the examples in the support set $S$ via the prototype vectors. An issue with this mechanism is it only employs the matching signals between the query example and the support examples for training. This can be acceptable for large datasets (e.g., in computer vision) where many examples can play the role of the query examples to provide sufficient training signals for the learning process. However, for EC, the available datasets are often small (e.g., the ACE 2005 dataset with only about a few thousands of annotated event mentions), making the sole reliance on the query examples for training signals less efficient. In other words, the few-shot learning framework might not be trained well with the limited data for the query matching for EC. Consequently, in this work, we propose to introduce more training signals for few-shot learning for EC by additionally exploiting the matching information among the examples in the support set themselves. In particular, as there are multiple examples (although only a few) per class/type in the support set, we select a subset of such examples for each type in $S$ and enforce the models to be able to match such the selected examples to their corresponding types in the remaining support set.



Formally, let $S_i = \{(s^1_i, a^1_i, t_i),\ldots, (s^{K}_i, a^{K}_i, t_i)\} \forall 1 \le i \le N$ so $S = S_1 \cup S_2 \ldots \cup S_N$. Let $Q$ be some integer that is less than $K$ (i.e., $1 \le Q < K$). For each type $t_i$, we randomly select $Q$ examples from $S_i$ (called the auxiliary query examples), forming the auxiliary query set $S^Q_i$ (i.e., $S^Q_i \subset S_i$, $|S^Q_i| = Q$). The remaining set of $S_i$ is then denoted by $S^S_i = S_i \setminus S^Q_i$. We unify the sets $S^S_i$ to constitute an auxiliary support set $S^S$ while the union of $S^Q_i$ serves as the auxiliary query set: $S^S = S^S_1 \cup S^S_2 \cup \ldots \cup S^S_N, S^Q= S^Q_1 \cup S^Q_2 \cup \ldots \cup S^Q_N$.


Given the auxiliary support set $S^S$, we seek to enhance the training signals for the few-shot models by matching the examples in the auxiliary query set $S^Q$ with $S^S$. Specifically, we first use the same networks in the instance encoder and prototypical modules to compute the auxiliary prototypes for the classes in $T$ of the auxiliary support set $S^S$. For each auxiliary example $z = (s_z, a_z, t_z) \in S^Q$ ($s_z$, $a_z$ and $t_z$ are the sentence, the trigger word position and the event type in $z$ respectively), we use the network in the classifier module to obtain the probability distribution $P(.|z,S^S)$ over the possible event types for $z$ based on the auxiliary support set $S^S$. Afterward, we enforce that the models can correctly predict the event types for all the examples in the auxiliary query sets $S^Q_i$ given the support set $S^S$ by introducing the auxiliary loss function:
\begin{equation}
\small
    L_{aux}(S) = -\sum^N_{i=1} \sum_{z=(s_z,a_z,t_i) \in S^Q_i} \log P(y = t_i|z,S^S)
    \label{eq:eq7}
\end{equation}

Eventually, the overall loss function to be optimized to train the models in this work is: $L(x,S) = L_{query}(x,S) + \lambda L_{aux}(S)$
where $\lambda$ is a trade-off parameter between the main loss function and the auxiliary loss function. For convenience, we call the training method with the auxiliary loss function for few shot learning in this section {\it LoLoss} (i.e., {\it leave-out loss}) in the following experiments.

\section{Experiments}
\subsection{Datasets and Hyper-Parameters}

We evaluate all the models in this study on the ACE 2005. ACE 2005 involves 33 event subtypes which are categorized into 8 event types: \textit{Business, Contact, Conflict, Justice, Life, Movement, Personnel, and Transaction}. The TAC KBP dataset, on the other hand, contains 38 event subtypes for  9 event types. Due to the larger numbers of the event subtypes, we will use the subtypes in these datasets as the classes for our few-shot learning problem.

As we want to maximize the numbers of examples in the training data, for each dataset (i.e., ACE 2005 or TAC KBP 2015), we choose the event subtypes in 4 event types that have the least number of examples in total and split at the ratio 1:1 into the test and development classes. Following this heuristics to select the classes, the event types used for training data in ACE 2005 involve {\it Business}, {\it Contact}, {\it Conflict}, and {\it Justice} while the event types for testing and development data are {\it Life}, {\it Movement}, {\it Personnel}, and {\it Transaction}. For TAC KBP 2015, the training classes include {\it Business}, {\it Contact}, {\it Conflict}, {\it Justice}, and {\it Manufacture} while the test and development classes consist of {\it Life}, {\it Movement}, {\it Personnel}, and {\it Transaction}. Finally, due to the intention to follow the prior work on few-shot learning with 10 examples per class in the support set and 5 examples per class in the query set for training \cite{gao2019hybrid}, we remove the examples of any subtypes whose have less than 15 examples in the training, test and development sets of the datasets. 


For the hyper-parameters, similar to the prior work \cite{gao2019hybrid}, we evaluate all the models using $N$-way $K$-shot FSL settings with $N,K \in \{5,10\}$. For training, we avoid feeding the same set of event subtypes in every batch to make training batches more diverse. Thus, following \cite{gao2019hybrid}, we sample 20 
event subtypes for each training batch while still keeping either 5 or 10 classes in the test time.

We initialize the word embeddings using the pre-trained GloVe embeddings with 300 dimensions. The word embeddings are updated during the training time as in \cite{Nguyen:15a}. We also randomly initialize the position embedding vectors with 50 dimensions. The other parameters are selected based on the development data of the datasets, leading to similar parameters for both ACE 2005 and TAC KBP 2015. In particular, the CNN encoder contains a single CNN layer with window size 3 and 250 filters. We manage to use this simple CNN encoder to have a fair comparison with the previous study \cite{gao2019hybrid}.
The Transformer encoder contains 2 layers with a context size of 512 and 10 heads in the attention mechanism. The number of examples per class in the auxiliary query sets $Q$ is set to 2 while the trade-off parameter $\lambda$ in the loss function is 0.1.


\subsection{Results}
Table \ref{tbl:ec-ace} shows the accuracy of the models (i.e., Matching, Proto, Proto+Att, and Relation) on the ACE 2005 test dataset, using the CNN encoder and Transformer encoder. There are several observations from the table. First, comparing the instance encoders, it is clear that the transformer encoder is significantly better than the CNN encoder across all the possible few-shot learning models and settings for EC. Second, comparing the few-shot learning models, the prototypical networks significantly outperform Matching and Relation with a large performance gap across all the settings. Among the prototypical networks, Proto+Att achieves better performance than Proto, thus confirming the benefits of the attention-based mechanism for the prototypical module. Third, comparing the pairs (5-way 5-shot vs 5-way 10-shot) and (10-way 5 shot vs 10 way 10 shot), we see that the performance of the models would be almost always better with larger $K$ (i.e., the number of examples per class in the support set) on different settings, consistent with the natural intuition about the benefit of having more examples for training. 

\input{tbl-ec-ace.tex}

Most importantly, we see that training the models with the LoLoss procedure would significantly improve the models' performance. This is true across different few-shot learning models, N-way K-shot settings, and encoder choices. The results clearly demonstrate the effectiveness of the proposed training procedure to exploit the matching information between examples in the support set for few-shot learning for EC. For simplicity, we only focus on the best few-shot learning models (i.e., the prototypical networks) and the Transformer encoder under 5-way 5-shot and 10-way 10-shot in the following analysis. Even though we show the results in fewer settings and models in table \ref{tbl:ec-tac} and \ref{tbl:noise}, the same trends are observed for the other models and settings as well.

Table \ref{tbl:ec-tac} additionally reports the accuracy of Transformer-based models on the TAC KBP 2015 dataset. As we can see from the table, most of our observations for the ACE 2005 dataset still hold for TAC KBP 2015, once again confirming the advantages of the proposed LoLoss technique in this work.

\input{tbl-ec-tac.tex}

\subsection{Robustness against noise}

\vspace{-0.1cm}

In this section, we seek to evaluate the robustness of the few-shot learning models against the possible noise in the training data. In particular, in each training episode where a set of examples is sampled for each type in $T$ to form the query set $Q$, we simulate the noisy data by randomly selecting a portion of the examples in $Q$ for label perturbation. Essentially, for each example in the selected subset of $Q$, we change its original label to another random one in $T$, making it a noisy example with an incorrect label. By varying the size of the selected portion in $Q$ for label perturbation, we can control the level of noise in the training process for FSL in EC.

\input{tbl-noise.tex}

Table \ref{tbl:noise} shows the accuracy of the Proto+Att model on the ACE 2005 test set that employs the Transformer encoder with or without the LoLoss training procedure for different noise rates. As we can see from the table, the introduction of noisy data would, in general, degrade the accuracy of the models (i.e., comparing the cells in Table \ref{tbl:noise} with the Proto+Att based model in Table \ref{tbl:ec-ace}). However, over different noise rates and N way K shot settings, the Proto+Att model trained with LoLoss is still always significantly better than those without LoLoss. The performance gap is substantial that is at least 4.5\% over different settings. In fact, we see that LoLoss can improve Proto+Att in the noisy setting (i.e., at least 4.5\%) more significantly than those in the setting without noisy data (i.e., at most 3.3\% on the 5 way 5 shot and 10 way 10 shot settings in Table \ref{tbl:ec-ace}). Such evidence further confirms the effectiveness and robustness against noisy data of LoLoss for few-shot learning due to its exploitation of the matching information between the examples in the support set.

\section{Conclusion}

In this paper, we perform the first study on few-shot learning for event classification. We investigate different metric learning methods for this problem, featuring the typical prototypical network framework with several choices for the instance encoder (i.e., CNN and Transformer). In addition, we propose a novel technique, called LoLoss, to train the few-shot learning models for EC based on the matching information for the examples in the support set. The proposed LoLoss technique is applied to different few-shot learning methods for different datasets and settings that altogether help to significantly improve the performance of the baseline models. In the future, we plan to examine LoLoss for few-shot learning for other NLP and vision problems (e.g., relation extraction, image classification).

\section*{Acknowledgments}


This research has been supported in part by Vingroup Innovation Foundation (VINIF) in project code VINIF.2019.DA18 and Adobe Research Gift. This research is also based upon work supported in part by the Office of the Director of National Intelligence (ODNI), Intelligence Advanced Research Projects Activity (IARPA), via IARPA Contract No. 2019-19051600006 under the Better Extraction from Text Towards Enhanced Retrieval (BETTER) Program. The views and conclusions contained herein are those of the authors and should not be interpreted as necessarily representing the official policies, either expressed or implied, of ODNI, IARPA, the Department of Defense, or the U.S. Government. The U.S. Government is authorized to reproduce and distribute reprints for governmental purposes notwithstanding any copyright annotation therein. This document does not contain technology or technical data controlled under either the U.S. International Traffic in Arms Regulations or the U.S. Export Administration Regulations.

\bibliographystyle{splncs04}
\bibliography{ref.bib}



\end{document}

%% file: tbl-ec-ace.tex
\begin{table}[h!]
\centering
\resizebox{\textwidth}{!}{
\begin{tabular}{|l|c|c|c|c|c|c|c|c| }
\hline
\multirow{2}{*}{\textbf{FSL Setting}} & \textbf{5 way}  & \textbf{5 way}   & \textbf{10 way} & \textbf{10 way}  & \textbf{5 way}  & \textbf{5 way}   & \textbf{10 way} & \textbf{10 way}  \\
               & \textbf{5 shot} & \textbf{10 shot} & \textbf{5 shot} & \textbf{10 shot} & \textbf{5 shot} & \textbf{10 shot} & \textbf{5 shot} & \textbf{10 shot} \\
\hline
 & \multicolumn{4}{|c|}{CNN Encoder} & \multicolumn{4}{|c|}{Transformer Encoder}  \\
\hline
Matching             & 45.81 & 49.01 & 30.41 & 35.66 & 71.83 & 76.51 & 61.2  & 66.79 \\ 
Matching+LoLoss    & \textbf{51.78} & \textbf{52.64} & \textbf{32.48} & \textbf{39.15} & \textbf{78.13} & \textbf{83.42} & \textbf{68.91} & \textbf{75.30} \\\hline
Proto                & 70.92 & 74.40 & 57.59 & 62.67 & 78.07 & 82.64 & 68.77 & 74.99 \\ 
Proto+LoLoss       & \textbf{76.98} & \textbf{82.19} & \textbf{66.92} & \textbf{73.63} & \textbf{81.27} & \textbf{86.20} & \textbf{73.07} & \textbf{79.63} \\
\hline
Proto+Att          & 72.26 & 74.22 & 57.28 &64.36 & 80.77 & 83.96 & 72.78 & 77.97 \\ 
Proto+Att+LoLoss & \textbf{76.93} & \textbf{75.59} & \textbf{67.54} & \textbf{66.70} & \textbf{83.38} & \textbf{87.20} & \textbf{76.03} & \textbf{81.79} \\
\hline
Relation          & 36.33 & 33.75 & 24.21 & 18.04 & 51.22 & 55.47 & 36.98 & 39.89 \\
Relation+LoLoss & \textbf{37.86} & \textbf{38.52} & \textbf{25.99} & \textbf{23.47} & \textbf{54.74} & \textbf{56.60} & \textbf{39.74} & \textbf{41.69} \\ 
\hline
\end{tabular}
}
\caption{Accuracy of event classification on ACE-2005 dataset. \textbf{+LoLoss} indicates the use of the auxiliary loss.}
\label{tbl:ec-ace}
\vspace{-0.2cm}
\end{table}

%% file: tbl-ec-tac.tex
\begin{table}[htbp]
\centering
\begin{tabular}{ |l | c | c | c | c |}
\hline
\textbf{Model} & \textbf{5 way 5 shot} & \textbf{10 way 10 shot}  \\
 \hline
Matching             & 72.78   & 65.55 \\
Matching+LoLoss    & \textbf{75.58} &   \textbf{68.53} \\ \hline
Proto                & 78.08 &   73.23 \\
Proto+LoLoss       & \textbf{78.88}   & \textbf{74.82} \\\hline
Proto+Att          & 75.35  & 71.28 \\
Proto+Att+LoLoss & \textbf{79.93}   & \textbf{76.37} \\\hline
Relation          & 50.97 & 34.91 \\
Relation+LoLoss & \textbf{51.65} & \textbf{35.13} \\
\hline
\end{tabular}
\caption{Accuracy of the models with the Transformer encoder on the TAC-KBP test dataset. \textbf{+LoLoss} indicates the use of the auxiliary loss.}
\label{tbl:ec-tac}
\vspace{-0.7cm}
\end{table}

%% file: tbl-noise.tex
\begin{table}[ht]
\centering
\addtolength{\belowcaptionskip}{-5mm}
\begin{tabular}{|c|l|c|c|}
\hline
\textbf{Noise rate} & \textbf{Model}  & \textbf{5 way 5 shot} & \textbf{10 way 10 shot} \\ 
\hline
\multirow{2}{*}{20\%} & Proto+Att  &  70.08 &  59.55 \\ 
      & Proto+Att+LoLoss  & \textbf{74.61} &  \textbf{64.66} \\ \hline
\multirow{2}{*}{30\%} & Proto+Att   & 67.38 &  57.08  \\ 
      & Proto+Att+LoLoss & \textbf{72.45}  & \textbf{62.65} \\ \hline
\multirow{2}{*}{50\%} & Proto+Att  & 60.50 & 50.67  \\ 
      & Proto+Att+LoLoss & \textbf{65.29} & \textbf{55.21} \\ 
      \hline
\end{tabular}
\caption{The accuracy on the ACE-2005 test set with different noise rates.}
\label{tbl:noise}
\end{table}